\documentclass[runningheads]{llncs}

\usepackage{eccv}

\usepackage{eccvabbrv}

\usepackage{graphicx}
\usepackage{booktabs}
\usepackage{stmaryrd}
\usepackage{textcomp, gensymb}

\usepackage[accsupp]{axessibility}  % Improves PDF readability for those with disabilities.

\usepackage{hyperref}

\usepackage{amsmath}
\usepackage{orcidlink}
\usepackage[accsupp]{axessibility}  % Improves PDF readability for those with disabilities.

\newcommand{\approach}{V3D\xspace}

\begin{document}

% \title{\approach: Video Diffusion Model as Multi-view Consistent Novel View Synthesizer} 
\title{\approach: Video Diffusion Models are Effective 3D Generators} 

\titlerunning{\approach}

\author{Zilong Chen\inst{1,2} \and
Yikai Wang\inst{1}$^\dagger$ \and
Feng Wang\inst{1} \and
Zhengyi Wang\inst{1,2} \and
Huaping Liu\inst{1}$^\dagger$
}

\authorrunning{Zilong Chen et al.}

\institute{Tsinghua University\\\email{chenzl22@mails.tsinghua.edu.cn} \and
ShengShu\\
}
\maketitle

\begin{figure}
    \centering
    \vspace{-9mm}
    \includegraphics[width=1.0\textwidth]{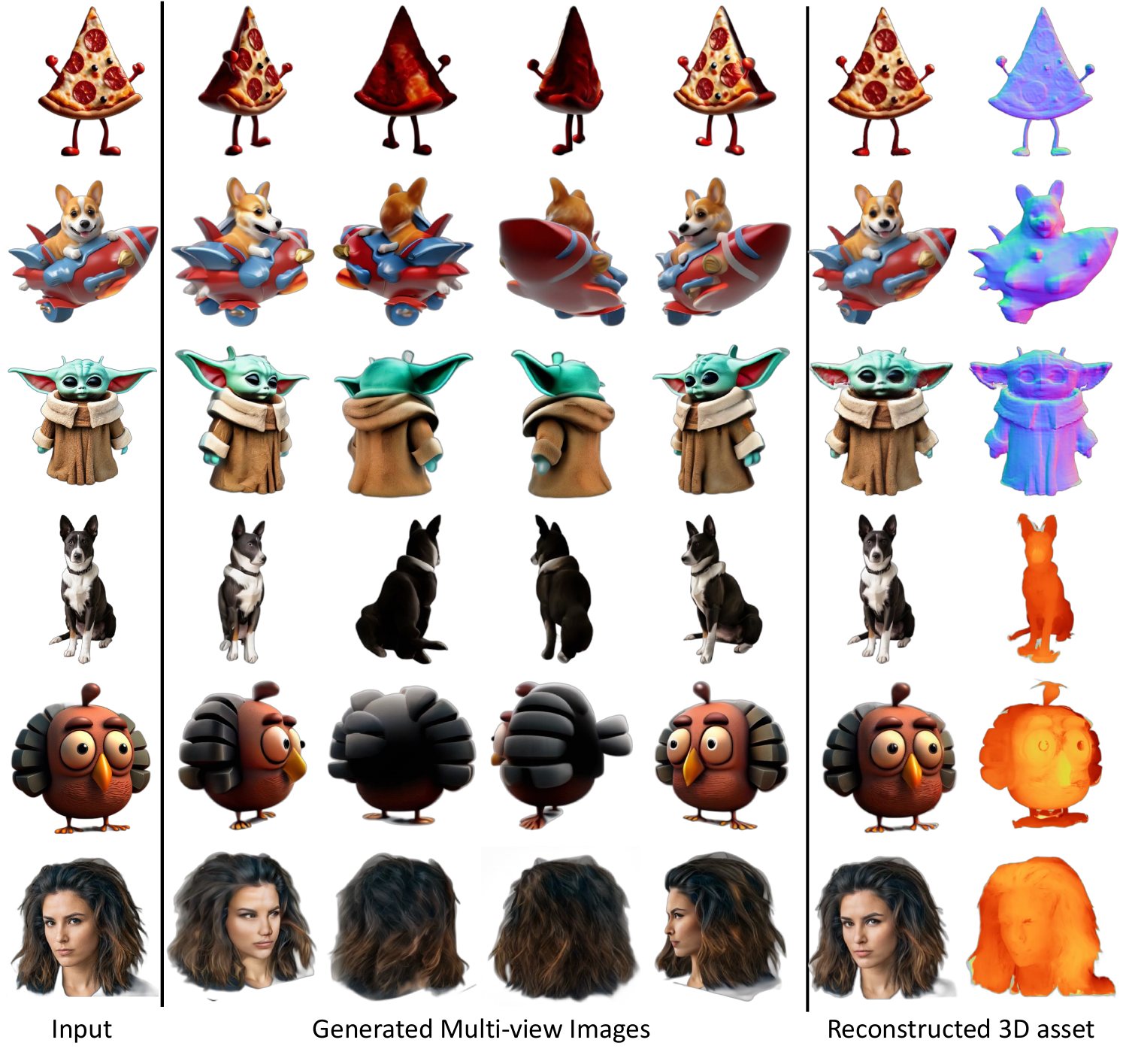}
    \vspace{-8mm}
    \caption{Delicate 3D assets generated by \approach. Our approach can generate high-fidelity 3D objects within 3 minutes.}
    \label{fig:opening}
    \vspace{-9mm}
\end{figure}

\let\thefootnote\relax\footnotetext{$^\dagger$Corresponding Authors.}
\begin{abstract}
  Automatic 3D generation has recently attracted widespread attention. 
  Recent methods have greatly accelerated the generation speed, but usually produce less-detailed objects due to limited model capacity or 3D data.
  Motivated by recent advancements in video diffusion models, we introduce \approach, which leverages the world simulation capacity of pre-trained video diffusion models to facilitate 3D generation. 
  To fully unleash the potential of video diffusion to perceive the 3D world, we further introduce geometrical consistency prior and extend the video diffusion model to a multi-view consistent 3D generator. 
  Benefiting from this, the state-of-the-art video diffusion model could be fine-tuned to generate $360\degree$ orbit frames surrounding an object given a single image. With our tailored reconstruction pipelines, we can generate high-quality meshes or 3D Gaussians within 3 minutes. 
  Furthermore, our method can be extended to scene-level novel view synthesis, achieving precise control over the camera path with sparse input views.
  Extensive experiments demonstrate the superior performance of the proposed approach, especially in terms of generation quality and multi-view consistency. Our code is available at \href{https://github.com/heheyas/V3D}{https://github.com/heheyas/V3D}.
\end{abstract}

\section{Introduction}
\label{sec:intro}
% 说明一下nvs这个任务和最近取得的进展和video diffusion
% Novel view synthesis 
% 以dreamfusion为代表的基于Score Distillation Sampling方法在3d生成领域取得了巨大的成功。但是，这类基于优化的方法速度较慢，并且有Mode collapse和Janus problem. 最近的方法提出了很多种其他路线来试图解决这个问题，包括直接将图片映射成三维表示\cite{}（例如triplane nerf或者3dgs），先生成多视角图片然后进行三维重建\cite{}。但是这些方法都存在一些问题，直接映射成三维表示难以利用数量级大很多的图片数据或者与训练的图像diffusion模型，而mv generation类的方法由于需要crossattn来保证多视角一致性，因此很难直接生成数量足够重建的多视角图片。基于这些观察，我们认为，video diffusion的特点可以极大的帮助3d生成和新视角合成。一方面，一组posed images可以被看做包含位恣信息的video，这样基于video diffusion可以使用数据丰富且质量更高的视频和图片数据和更大模型的与训练信息，另一方面，video diffusion的结构使得模型可以再有限的计算资源里生成足够的多视角图片，后续可通过直接进行三位重建来得到对应的三维模型。
% Diffusion model based image generation has achieved 
With the advancement of diffusion-based image generation, automatic 3D generation has also made remarkable progress recently.
Approaches based on Score Distillation Sampling~\cite{dreamfusion} have achieved tremendous success in the field of 3D generation by extracting priors directly from text-to-image diffusion models~\cite{prolificdreamer, magic3d, fantasia3d, latentnerf, chen2024gsgen}.
However, such optimization-based methods are relatively slow and suffer from mode collapse~\cite{wang2023esd, prolificdreamer} and the Janus problem~\cite{perpneg, seo2023let}.

Subsequent efforts have proposed various alternative pathways to address these issues, mainly including directly mapping images into 3D representations \cite{zou2023triplane, tang2024lgm, LRM, PF-LRM, shap_e, pointe} (such as Triplane NeRF~\cite{Chan2021eg3d} or 3DGS~\cite{kerbl3Dgaussians}), or first generating consistent multi-view images then conducting 3D reconstruction to obtain the underlying 3D model \cite{long2023wonder3d, liu2023syncdreamer, liu2023one2345}. 
With the increase in 3D data size and model capacity, these methods have reduced the generation time from hours to minutes.
% However, these methods encounter several challenges.
However, the training process of these models still exhibits certain limitations;
% Directly mapping into 3D representations can only be trained on 3D datasets, making it difficult to leverage higher-quality, larger-scale image data and pre-trained image diffusion models. As a result, the accuracy of these models is often limited.
% Directly mapping into 3D representations can only be trained on 3D datasets, limiting the utilization of higher-quality, larger-scale image data and pre-trained image diffusion models, thus often constraining the accuracy of these models.
Methods that directly map to 3D representations require training on 3D datasets, making it impractical to leverage higher-quality image data and large-scale pre-trained image diffusion models.
Meanwhile, approaches based on multi-view generation face difficulties in directly generating a sufficient number of images for reconstruction due to the requirement of memory-consuming cross-attention or intermediate 3D structure to ensure consistency across views. 

Recently, video diffusion models have attracted significant attention due to their remarkable ability to generate intricate scenes and complex dynamics with great spatio-temporal consistency~\cite{Blattmann2023StableVD, blattmann2023videoldm, sora, guo2023animatediff, chen2024videocrafter2, MakePixelsDance, bar-tal2024lumiere, girdhar2023emu}. Many videos incorporate changes in perspectives which provide natural observations of 3D objects from different views~\cite{li2024sora}, enabling the diffusion models to perceive the physical 3D world~\cite{sora}. 
Current multi-view generation methods are based on image diffusion models, which can only generate a few views (less than 6). In contrast, video diffusion models can generate hundreds of frames, which is adequate for downstream 3D tasks.
These advancements give us the possibility to generate consistent multi-view images using video diffusion models and then reconstruct underlying 3D assets to achieve high-quality 3D generation. 
% However, 3D reconstruction requires pixel-level consistency which is difficult for current video diffusion models. 
However, reconstruction from multi-views imposes significant demands on consistency which is difficult for current video diffusion models. 
In this paper, we aim to fully unleash the potential of video diffusion models to perceive the 3D world and inject the geometrical consistency prior to make the diffusion model an effective multi-view generator.
In light of this, we propose \approach, a novel approach for 3D generation with video diffusion models. Specifically, we propose to fine-tune video diffusion models on 3D datasets with additional conditions to generate novel views in both object-centric and scene-level scenarios. For object-centric 3D generation, we fine-tune the base video diffusion model on $360\degree$ orbit videos of synthetic 3D objects with the front view as a condition.
We find that our fine-tuned model can generate reliable multi-views for 3D reconstruction. However, there still exists inconsistency in some details which may diminish the reconstruction quality. To address this, we adopt perceptual loss instead of pixel-wise loss as the reconstruction objective. We employ Gaussian Splatting as the 3D representation due to its reconstruction and rendering efficiency. To further accelerate the reconstruction, we propose a space-carving initialization method for Gaussians, which involves unprojecting multi-view object masks back to 3D space to roughly position the Gaussians on the object's surface, eliminating the invalid optimizations on empty spaces.
Considering the demands of real-world applications, we also propose a novel mesh extraction method that initially extracts the surface using SDF and subsequently refines the appearance on generated views with image-level loss.
Furthermore, we demonstrate the effectiveness of our approach in scene-level 3D generation. To achieve precise camera path control and accommodate multi-view input, we integrate a PixelNeRF encoder to enhance the video diffusion model with robust 3D signals. Combined with auxiliary loss and fine-tuning on real-world posed videos, we extend the video diffusion model to generate novel views on an arbitrary camera path given sparse view input.
% As a result of these techniques, our approach can generate high-fidelity 3D assets given a single view of an object and synthesize novel views on an arbitrary camera path given several posed reference images in a scene. 
% \red{\cref{fig:opening} demonstrates delicate 3D assets generated with \approach.} 
% pixelnerf + 重建 （初始化 Spcae Carving） + 提mesh
Our contributions are:
% \vspace{-2mm}
\begin{itemize}
    % \item We propose \approach, a novel approach for both object-centric and scene-level novel view synthesis. By tailoring and fine-tuning the base video diffusion model on synthetic and real-world datasets, we explored the capability of video diffusion as a multi-view consistent novel view synthesizer.
    % \item We propose \approach, a novel approach for object-centric and scene-level 3D generation with video diffusion models. By incorporating 3D priors, we explore the potential of video diffusion models as effective 3D generators. 
    \item We propose \approach, a framework that first employs a video diffusion model to generate consistent multi-view frames, then reconstructs the underlying 3D content. 
    % We successfully fine-tune the video diffusion model on 3D datasets to boost the ability of geometrical consistency. Our framework is universal and can be applied to both object and scene generation.
    By fine-tuning on 3D datasets, we successfully boost the ability of geometrical consistency of video diffusion models. Our framework is generally applicable to both object and scene generation.
    % \item With the aid of 3D Gaussian Splatting that is capable of incorporating perceptual loss, we propose a novel approach for image-to-3D which generates dense novel views with a fine-tuned video diffusion model and then robustly reconstructs the underlying 3D object.
    \item We design a reconstruction pipeline tailored for video diffusion outputs to generate high-quality 3D Gaussians or textured meshes. With our designed efficient initialization and refinement, \approach can reconstruct delicate 3D assets within 3 minutes.
    % \item By introducing a PixelNeRF-based encoder, we further extend the video diffusion model to generate novel views on an arbitrary camera path.
    % \item We evaluate \approach in various scenarios. Experiments demonstrate that our approach can generate multi-view consistent novel views with high-resolution and exceptional fidelity.
    \item We conduct experiments to validate the effectiveness of \approach. Both object-centric and scene-level experiments show that our method achieves state-of-the-art performance in reconstruction quality and multi-view consistency.
    % Besides, we extend our generation framework to scene-level and achieves state-of-the-art performances.
    % \item rubbish
\end{itemize}

\section{Related Work}

% 不要了
% \subsection{Video Diffusion Models}
% Video diffusion models have garnered significant attention in recent years due to their ability to generate high-quality video sequences. These models leverage the principles of diffusion processes to iteratively update pixel values in a sequence of frames, resulting in smooth and coherent temporal dynamics. Early work focused on 
% \newline
% The most relevant work with us is Stable Video Diffusion \cite{Blattmann2023StableVD}, which 

\subsection{3D Generation}
% 相关工作
% 多视角生成
% 为了从单个图片生成3d模型，很多方法选择了把image to image模型拓展成多视角生成模型。zero123最早提出了将sd finetune成mv diffusion的方法，并且将camera pose当做额外的condition来生成任意视角的图片；
%mvdream，image dream
%wonder3d
%direct2.5
%syncdreamer
%sweetdreamer
Early efforts in 3D generation focus on per-scene optimization methods based on CLIP~\cite{CLIP, sanghi2021clip, jain2021dreamfields, text2mesh, TANGO, clipnerf, khalid2022clipmesh}. DreamFusion~\cite{dreamfusion}, the pioneer work, leverages stronger diffusion prior by introducing Score Distillation Sampling that minimizes the difference between rendered images from the underlying 3D assets and the diffusion prior. Subsequent methods have further achieved substantial improvements of this paradigm in quality~\cite{prolificdreamer, fantasia3d, magic3d, hifa}, optimization speed~\cite{tang2023dreamgaussian, latentnerf}, alleviating the Janus problem~\cite{perpneg, seo2023let, chen2024gsgen, shi2023MVDream, sweetdreamer}, and have expanded their applications to generative tasks such as editing~\cite{GaussianEditor, DDS, dreameditor, ednerf, wang2023animatabledreamer}, texture generation~\cite{zeng2023paint3d, fantasia3d, latentnerf}, and single image-to-3D~\cite{wang2023imagedream, Magic123, tang2023makeit3d, sun2023dreamcraft3d}. Despite great success has been achieved, optimization-based methods still suffer from slow generation speed and low success rates. To overcome these challenges, researchers have explored some non-optimization paradigms. One stream involves first generating consistent multi-view images, and then utilizing reconstruction methods to obtain the corresponding 3D model~\cite{liu2023one2345, liu2023one2345++, tang2023MVDiffusion, tang2024mvdiffusionpp}. SyncDreamer~\cite{liu2023syncdreamer} combines explicit voxels and 3D convolution with a diffusion model to enhance the consistency of generated multi-views. Wonder3D~\cite{long2023wonder3d} fine-tunes image diffusion models to generate sparse views with corresponding normal maps and utilizes a NeuS~\cite{Wang2021NeuS} to reconstruct the underlying geometry. Direct2.5~\cite{Lu2023Direct25DT} adopts a remeshing-based technique to recover geometry from generated sparse normal maps. Another line of research involves directly mapping sparse-views into 3D representations~\cite{szymanowicz23splatter, liu2023unidream, szymanowicz23viewset_diffusion, DMV3D, charatan2023pixelsplat, yariv2023mosaicsdf, Wang2023rodin, xu2024agg, qian2024atom, meshgpt}. LRM~\cite{LRM}, PF-LRM~\cite{PF-LRM}, and Instant3D~\cite{instant3d} adopt a huge transformer to directly predict triplanes from single or sparse views. TriplaneGaussian~\cite{zou2023triplane} and LGM~\cite{tang2024lgm} instead map sparse views into more memory-efficient 3D Gaussian Splatting which supports much higher resolution for supervision. Our concurrent work IM-3D~\cite{MelasKyriazi2024IM3DIM} also explores the capability of video diffusion models in object-centric multi-view generation, we further extend this paradigm in scene-level novel view synthesis and achieve better performance.

\subsection{Generative Models for Novel View Synthesis}
While NeRF~\cite{nerf} and 3D Gaussian Splatting~\cite{kerbl3Dgaussians} have shown impressive performance in novel view synthesis with a sufficient number of inputs, reconstruction from sparse views requires additional priors due to the incomplete information provided. Early work mainly focuses on using regression-based~\cite{ibrnet, pixelnerf, SRT, Chen2021MVSNeRFFG, Flynn2015DeepSL, Henzler2021UnsupervisedLO, Trevithick2020GRFLA, Tucker2020SingleViewVS, Zhou2018StereoM} or GAN-based~\cite{Chan2021eg3d, chanmonteiro2020pi-GAN, Gadelha20163DSI, Gao2022GET3DAG, Henzler2018EscapingPC, NguyenPhuoc2019HoloGANUL, Niemeyer2020GIRAFFERS, Schwarz2022VoxGRAFF3, Zhu2018VisualON} methods for generalizable scene reconstruction. However, due to the lack of high-quality data and the limited model capacity, the generated novel views are often blurry and exhibit poor generalization ability. Subsequent work further integrates diffusion priors to achieve better scene-level novel view synthesis~\cite{Watson2022NovelVS, karnewar2023holodiffusion}. SparseFusion~\cite{sparsefusion} proposes view-conditioned diffusion with epipolar feature transformer to synthesize novel views with sparse inputs. GeNVS~\cite{chan2023genvs} utilizes a PixelNeRF encoder to incorporate geometry information and exploits an auto-aggressive generation strategy to enhance multi-view consistency. ZeroNVS~\cite{zeronvs} extends Zero-1-to-3~\cite{liu2023_zero1to3} to unbounded scenes and resolves the scale ambiguity with better camera pose conditioning. ReconFusion~\cite{wu2023reconfusion} proposes to fine-tune image diffusion models to condition on multi-view input and utilizes a novel sample loss to reconstruct the 3D scene from sparse posed inputs. We draw inspiration from ReconFusion and GeNVS, adopting PixelNeRF-based conditioner to accommodate any number of input images and provide precise camera path control in scene-level generation.

\section{Approach}
\subsection{Overall Structure}
\begin{figure}[t]
    \centering
    \includegraphics[width=0.9\textwidth]{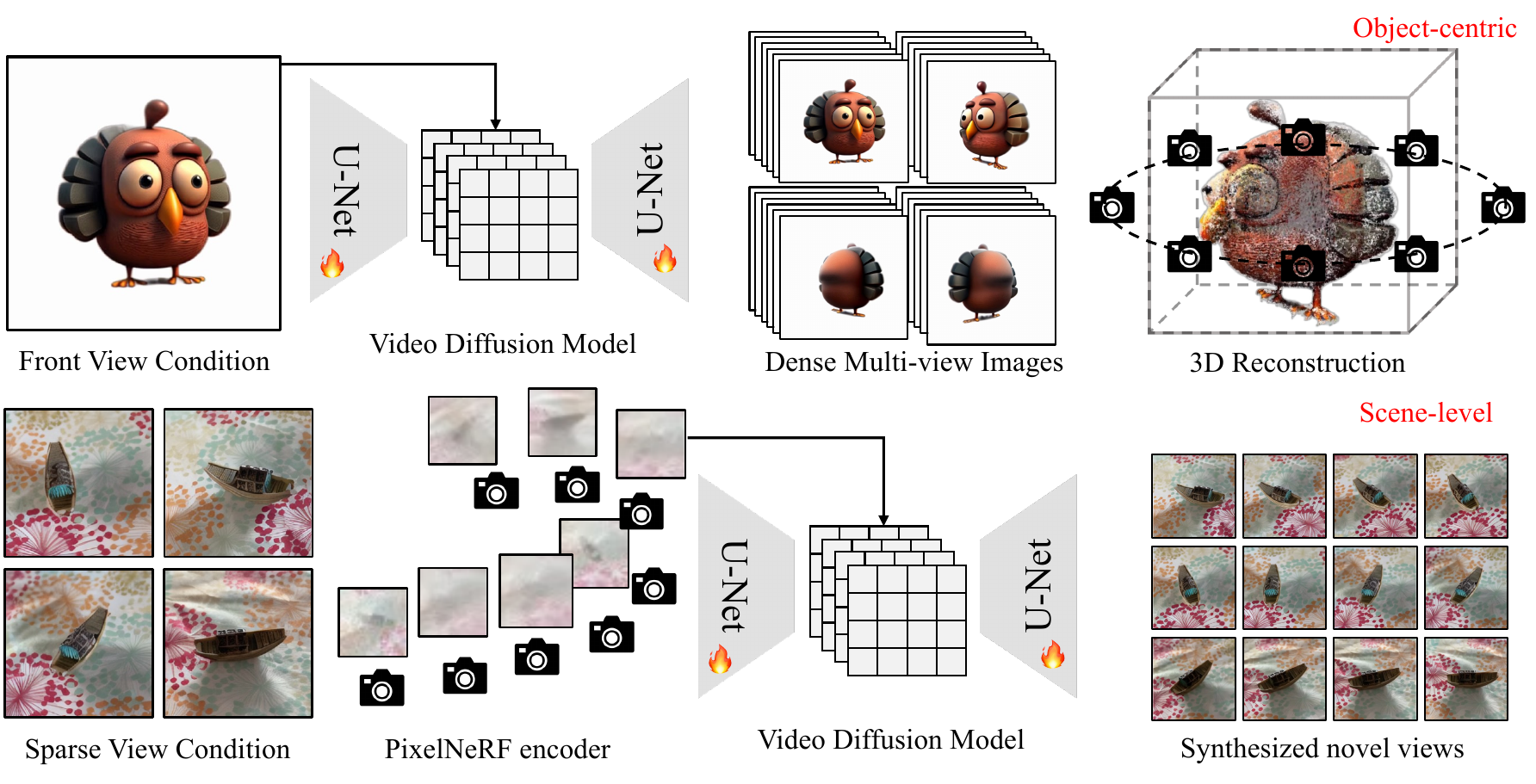}
    \caption{Overview of the proposed \approach.}
    \label{fig:overall}
    \vspace{-5mm}
\end{figure}
% 如图所示，\approach是一个两阶段模型。给一个图片，\approach首先使用微调之后的video diffusion模型来生成dense view的图片。具体来说，我们将物体的多视角图片序列当做一个video，并基于svd-xt，一个开源的image-to-video模型，在三维数据集上进行微调。微调之后的模型可以根据给定的图片作为条件，生成对应的dense multi-view images。在得到多视角图片，我们使用gaussian splatting来从dense view中重建出三维模型，并且结合video。
% 除此之外，结合pixelnerf encoder的方式加入camera path condition，我们进一步将video diffusion model拓展到了场景级的新视角合成上。

% 这一段要重写
% As illustrated in \cref{fig:overall}, \approach is a two-stage model that consists of dense view generation and 3D reconstruction. Given an image, \approach first utilizes a fine-tuned video diffusion model to generate dense views of the underlying 3D asset. 
As demonstrated in \cref{fig:overall}, the core insight of \approach lies in conceptualizing dense multi-view synthesis as video generation, thus leveraging the structure and powerful priors of the large-scale pre-trained video diffusion model to facilitate consistent multi-view generation.
% \red{Concretely, by injecting 3D knowledge via 3D video data fine-tuning, we expand the capability of video generation models to encompass object-centric and scene-level view synthesis.}
For object-centric image-to-3D, we fine-tune the base video diffusion model on $360\degree$ orbit videos of synthetic 3D objects rendered on fixed circular camera poses (\cref{sec:object-finetune}) and design a reconstruction and mesh extraction pipeline that is tailored for generated multi-views (\cref{sec:reconstruction}). 
For scene-level novel view synthesis, we enhance the base video diffusion model by incorporating a PixelNeRF encoder, enabling it to precisely control camera poses of generated frames and seamlessly adapt to any number of input images (\cref{sec:scene-nvs}).
The detailed \approach methodology is presented as follows.
\subsection{Dense View Prediction as Video Generation}
\label{sec:object-finetune}
% 以往的多视角生成模型往往基于image diffusion model进行拓展，使用multi-view cross attention(Wonder3D)或者3D voxel反投影结合3D CNN来增强生成结果的多视角一致性(SyncDreamer)。前者由于使用了复杂度为n^2的cross attn，因此视角的数量往往比较少，而后者由于使用了显示的3d voxel，因此最后图像的分辨率比较低。\approach的多视角合成采取了完全不同的方法。受到SVD的启发，我们将多视角生成看作视频生成，从而可以很好的利用SVD在大规模视频数据上预训练先验，并且利用video diffusion模型的时空一致性来提升多视角一致性。具体来说，
Previous multi-view generation models~\cite{CascadeZero123, weng2023consistent123, tang2023MVDiffusion} often extend from image diffusion models, incorporating multi-view cross attention (e.g. Wonder3D~\cite{long2023wonder3d} and Zero123-XL~\cite{shi2023zero123plus}) or intermediate 3D representations (e.g. SyncDreamer~\cite{liu2023syncdreamer} with 3D voxels and Viewset Diffusion~\cite{szymanowicz23viewset_diffusion} and DMV3D~\cite{DMV3D} with triplane NeRF) to enhance the consistency of generated images across different views. 
The former tends to involve a limited number of views due to the use of cross-attention with quadratic complexity, whereas the latter often results in low image resolution due to memory-consuming 3D representations and volume rendering.
To overcome these drawbacks, \approach adopts a fundamentally distinct approach to generate dense multi-view images given a single view. 
Inspired by Stable Video Diffusion (SVD)~\cite{Blattmann2023StableVD}, we interpret continuous multi-view images rotating around an object as a video, thereby treating multi-view generation as a form of image-to-video generation with the front view as the condition.
% \red{The advantage of this approach lies in its ability to leverage the structure of video diffusion local attention to significantly increase the quantity and resolution of generated images.}
% \red{Moreover, it allows us to effectively utilize the pre-training information of video diffusion to address the challenges posed by the limited quantity and quality of 3D data.}
% The advantage of this is that it leverages the understanding of the 3D world provided by large-scale pre-trained video diffusion models to overcome the scarcity of 3D data, and utilizes the network architecture of video diffusion models to efficiently generate a sufficient number of multi-view images. 
This approach capitalizes on the comprehensive understanding of the 3D world offered by large-scale pre-trained video diffusion models to address the scarcity of 3D data and harnesses the inherent network architecture of video diffusion models to effectively produce an adequate number of multi-view images.
% \red{For object-centric image-to-3D, we [lack of one connection sentence]}
\subsubsection{Conditioning.}
Similar to SVD, we provide high-level semantic information of the front view by injecting the corresponding CLIP embedding into the diffusion U-Net via cross-attention and prompt the diffusion model low-level information by concatenating its latent along the channel dimension of the input.
To better adapt to image-to-3D, we removed irrelevant conditions including motion bucket ID and FPS ID, and opted not to condition on elevation angle. Instead, we randomly rotate the object in elevation to make the generation model compatible with inputs with non-zero elevation.
Our modification to the conditioning arises from experimental evidence suggesting that it is not feasible to perfectly constrain the elevation of generated multi-view images to a specified angle. The misalignment between the camera poses of the video and those used for reconstruction could lead to significant damage to the final outcomes.
% Our motivation behind this modification stems from experimental observations indicating that it is not feasible to perfectly fix the elevation of generated multi-view images to a specified angle through conditioning and the misalignment between camera poses of the video and those used for reconstruction could cause catastrophic damage to the final outcomes.
% Besides, distinct from Stable Video Diffusion, we remove the motion bucket and FPS conditions due to it being unnecessary in our setting. 
% Moreover, we opted not to incorporate elevation as input to the diffusion model, as we discovered that this approach does not perfectly constrain the camera path to the desired elevation. Consequently, the resulting misalignment between the camera path used for video generation and the one used for reconstruction can lead to catastrophic damage to the reconstruction outcomes.
% \subsubsection{Loss.}
% 我们使用了central guidance scale来sample这个video diffusion model

\subsubsection{Data.}
We fine-tune the base video diffusion model on the Objaverse dataset~\cite{objaverse} for object-centric image-to-3D.
It is well-established that the Objaverse dataset contains quite many low-quality 3D objects~\cite{long2023wonder3d, liu2023one2345++} which may significantly degenerate the model performance~\cite{liu2023_zero1to3}. 
We, therefore, start by manually curating a subset with higher quality, comprising approximately 290k synthetic 3D triangle meshes. We then construct a $360\degree$ orbit video dataset by rendering synthetic 3D objects in this filtered subset on $N=18$ fixed camera poses. Specifically, for each object, we normalize the mesh to unit length, fix the camera distance to 2, set the elevation to 0, and uniformly move the camera on the azimuth $N$ times (from 0 to $2\pi$) to generate a smooth orbit video at 512 resolution.

\subsubsection{Training.}
% 还有learning rate之类的一大堆屁话没说，这部分可以考虑全都放到附录
We follow SVD to adopt the commonly used EDM~\cite{Karras2022edm} framework with the simplified diffusion loss for fine-tuning. 
For the noise distribution used in training, we inspired by \cite{Lin2023CommonDN, shi2023zero123plus}, adopt a relatively large noise distribution with $P_{\text{mean}}=1.5$ and $P_{\text{std}}=2.0$.
To enable classifier-free guidance, we randomly set the latent and the CLIP embedding of the front view to zero independently with a probability of 20\%.
To speed up training and save GPU VRAM, we preprocess and store the latent and CLIP embedding of all video frames beforehand. During training, we load these tensors directly from the disk instead of computing them on the fly. We fine-tuned the base diffusion model for 22.5k steps, taking about one day. More detailed training settings are provided in the supplemental material.

\subsection{Robust 3D Reconstruction and Mesh Extraction}
\label{sec:reconstruction}
% 在使用\approach得到一个物体的多视角图片后，下一步是使用这些dense视角来恢复出the underlying 3D object. 尽管现在有很多很好的方法可以进行重建，我们选择了3D Gaussian Splatting，最近在重建上的state-of-the-art.
% 尽管\approach得到的多视角视频已经足够好了，但是仍然不能保证严格意义上的pixel-level的一致性，在这种情况下，直接使用L2 loss进行重建很容易收到图像之间不一致的影响，产生floater等artifact.
After obtaining dense views of an object using the fine-tuned video diffusion model, our next step is to reconstruct the underlying 3D object from these views. While the obtained multi-view images are of high resolution and satisfactory consistency, it's challenging to ensure strict pixel-to-pixel alignment between different views. In this case, directly applying pixel-wise loss for 3D reconstruction often leads to artifacts such as floaters or blurry textures~\cite{barron2021mip, park2023camp, lin2021barf}. In response, we propose using image-level perceptual loss to resolve inconsistencies among multiple input images. To support image-level loss, we opt for 3D Gaussian Splatting due to its fast training speed and great memory efficiency that enables rendering the full image with minimal cost. We adopt LPIPS~\cite{LPIPS} as the perceptual loss and combine it with the original loss for 3DGS, i.e.
\begin{equation}
    % \mathcal{L}_{\text{recon}}(I, I^{\text{gt})=\text{MSE}(I, I^{\text{gt}) + \lambda_{s}\text{D-SSIM}(I, I^{\text{gt}}) + \lambda_{l}\text{LPIPS}(I, I^{\text{gt}})
    \mathcal{L}_{\text{recon}}(I, I^{\text{gt}}) = \text{MSE}(I, I^{\text{gt}}) + \lambda_{s}\text{D-SSIM}(I, I^{\text{gt}}) + \lambda_{l}\text{LPIPS}(I, I^{\text{gt}})
\end{equation}
where $I$ and $I^{\text{gt}}$ represent the rendered images and the ground truth images (our generated views), $\lambda_s$ and $\lambda_l$ refer to the loss weights.

\subsubsection{Initialization.} Initialization is crucial for 3D Gaussian Splatting to achieve promising results~\cite{kerbl3Dgaussians, zhu2023FSGS}. 
% \red{The original 3DGS adopts Gaussian-blob initialization for the reconstruction of bounded objects and removing xx Gaussians by reset and threshold on opacities.}
Due to no sparse point cloud available in object-centric reconstruction, the original 3DGS initializes the Gaussians with Gaussian-blob and removes unnecessary Gaussians by resetting opacities.
This requires many optimization iterations to eliminate the floaters, and artifacts may occur when the input images are not perfectly consistent.
To accelerate the reconstruction, we propose to initialize the Gaussians by space carving~\cite{Kutulakos1999ATO, Lu2023Direct25DT}. Concretely, we first segment the foreground of the generated frames with an off-the-shelf background removal model to obtain an object mask for each view. An occupancy grid is then established by projecting voxels onto image planes to determine whether the projected pixel belongs to the object and aggregating the results of all projections under all views. Finally, we exploit marching cubes~\cite{Lorensen1987MarchingCA} to obtain a surface according to this occupancy grid and uniformly sample $N_{\text{init}}$ points on the surface as the initializing positions of the Gaussians.

\subsubsection{Mesh Extraction.} 
For the demand of real-world applications, we also propose a mesh extraction pipeline for generated views.
Similar to \cite{long2023wonder3d, tang2024mvdiffusionpp, liu2023syncdreamer}, we adopt NeuS~\cite{Wang2021NeuS} with multi-resolution hash grid~\cite{mueller2022instant} for fast surface reconstruction. 
While \approach can generate a considerable number of views, it remains modest compared to the typical use cases of NeuS, which commonly involve more than 40 posed images~\cite{Wang2021NeuS, Long2022SparseNeuSFG}. Therefore we adopt a normal smooth loss and a sparse regularization loss to ensure a reasonable geometry.
Besides, since our generated images are not perfectly consistent and it is not feasible for NeuS to exploit image-level loss due to the high render cost, the textures of the mesh extracted by NeuS are often blurry.
% To address this, we further keep the geometry of the NeuS extracted mesh unchanged and refine its texture with generated multi-views using LPIPS loss
To address this, we further refine the texture of the NeuS extracted mesh with generated multi-views using LPIPS loss while keeping the geometry unchanged, which greatly reduces the rendering cost and enhances the quality of final outputs. Owing to the great efficiency of differentiable mesh rendering~\cite{Laine2020diffrast}, this refinement process can be completed within 15 seconds.

\begin{figure}[!t]
    \centering
    \includegraphics[width=1.0\textwidth]{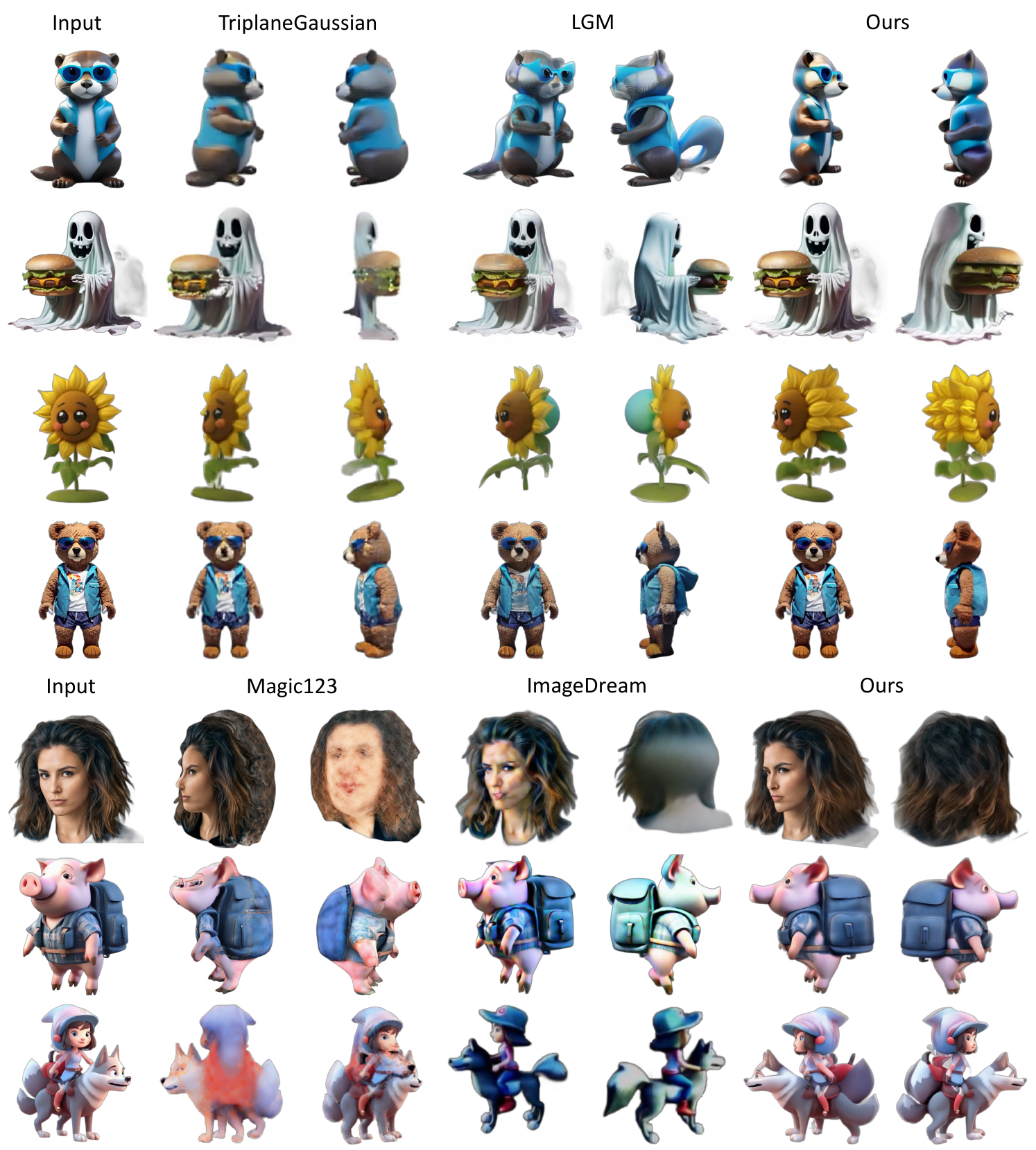}
    \caption{Qualitative comparison between the proposed \approach and state-of-the-art 3DGS-based image-to-3D methods (LGM~\cite{tang2024lgm} and TriplaneGaussian~\cite{zou2023triplane}) and optimization-based methods (Magic123~\cite{Magic123} and ImageDream~\cite{wang2023imagedream}). The corresponding $360\degree$ videos are demonstrated in the supplemental video.}
    \label{fig:gs_comparison}
\end{figure}

\subsection{Scene-level Novel View Synthesis}
\label{sec:scene-nvs}
% 除了object-centric的生成，我们进一步把video diffusion扩展到了场景级的新视角合成上。和生成中心化的物体不同的是，新视角合成需要生成给定camera pose path上的图像，而不只是固定视角的图片。之前的方法zero123使用了加入camera embedding的方式给video diffusion加入camera pose的条件，但是这种方式没有显式的利用3D信息，因此很难保证多个视角的一致性~\cite{}. 为了更好的使video diffusion感受到3D信息，我们受到GenVS和ReconFusion的启发，在video diffusion的基础上加入了一个pixelnerf feature encoder module作为条件来帮助更好的感知3d信息. 具体来说，给定一组多视角图片和目标的camera path，我们使用pixelnerf来渲染目标camera path对应的feature map
In addition to object-centric generation, we further extend video diffusion to scene-level novel view synthesis. Unlike generating views of objects, novel view synthesis entails generating images along a given camera path, which requires precise control over camera poses and accommodating multiple input images. 
\subsubsection{Conditioning.}
Previous approaches, such as Zero-1-to-3~\cite{liu2023_zero1to3}, incorporate camera poses into image diffusion by introducing camera embeddings. However, this method did not explicitly utilize 3D information, thus making it challenging to ensure consistency across multiple views and precise control over camera pose~\cite{liu2023one2345, wu2023reconfusion}. 
To better incorporate 3D prior into video diffusion and precisely control the poses of generated frames, we draw inspiration from GeNVS~\cite{chan2023genvs} and ReconFusion~\cite{wu2023reconfusion}, integrating a PixelNeRF~\cite{pixelnerf} feature encoder into video diffusion models.
Specifically, given a set of posed images as condition $\pi_{\text{cond}}, I_{\text{cond}}=\{(I_i, \pi_i)\}_i$, we adopt a PixelNeRF to render the feature map of the target camera poses $\{\pi\}$ by:
\begin{equation}
    f=\text{PixelNeRF}(\{\pi\}|\pi_{\text{cond}}, I_{\text{cond}}),
\end{equation}
where $\{\pi\}$ refers to the camera poses of the frames we want to generate. 
We then concatenate this feature map to the input of the U-Net to explicitly encode relative camera pose information. This approach seamlessly supports an arbitrary number of images as input. Other conditionings are similar to those of object-centric generation, except that the number of CLIP image embedding has been increased from one to multiple.

\subsubsection{Data.}
For scene-level novel view synthesis, we fine-tune the base video diffusion model on MVImgNet~\cite{yu2023mvimgnet} which consists of 219k posed real-world video clips. To make the videos from MVImgNet compatible with the input size of the video diffusion, we obtain the object masks with an off-the-shelf background removal tool, then crop, recenter the image, and adjust the principal points and focal lengths according to the foreground mask. 

\subsubsection{Training.} 
% \subsubsection{Loss.} 
Except for the simplified diffusion loss used in video diffusion models, we follow ReconFusion~\cite{wu2023reconfusion} to regulate the parameters of the PixelNeRF encoder with a down-sampled photometric loss:
\begin{equation}
    \mathcal{L}_{\text{PixelNeRF}}=\mathbb{E}\lVert f_{\text{RGB}}(\pi|x_{\text{cond}}, \pi_{\text{cond}})-x_{\downarrow} \rVert
\end{equation}
where $f_{\text{RGB}}$ refers to another color branch of the PixelNeRF encoder that predicts a downsampled image and $x_{\downarrow}$ refers to the downsampled target frames. We experimentally found this loss crucial to avoid local minima, as illustrated in~\cite{wu2023reconfusion}.
We follow InstructPix2Pix~\cite{brooks2022instructpix2pix}, initializing the newly added input channels for the PixelNeRF feature in the first layer of the diffusion U-Net with zero to best keep the pre-trained priors. For input views conditioning, we randomly select 1 to 4 images as input and set the conditions to zeros independently with a probability of 20\%. We present more training details of the scene-level model in the supplemental material.

\section{Experiments}
\label{sec:experiments}
In this section, we present our experiments on validating the effectiveness of the proposed approach in both object-centric and scene-level 3D generation. Specifically, we quantitatively and qualitatively compare \approach with previous state-of-the-art methods in image-to-3D and novel view synthesis. Additionally, we conduct several ablation studies to evaluate the importance of model designs, including noise distribution, number of fine-tuning steps, pre-training prior, and camera conditioning. The detailed results are shown as follows.

\begin{figure}[!t]
    \centering
    \includegraphics[width=1.0\textwidth]{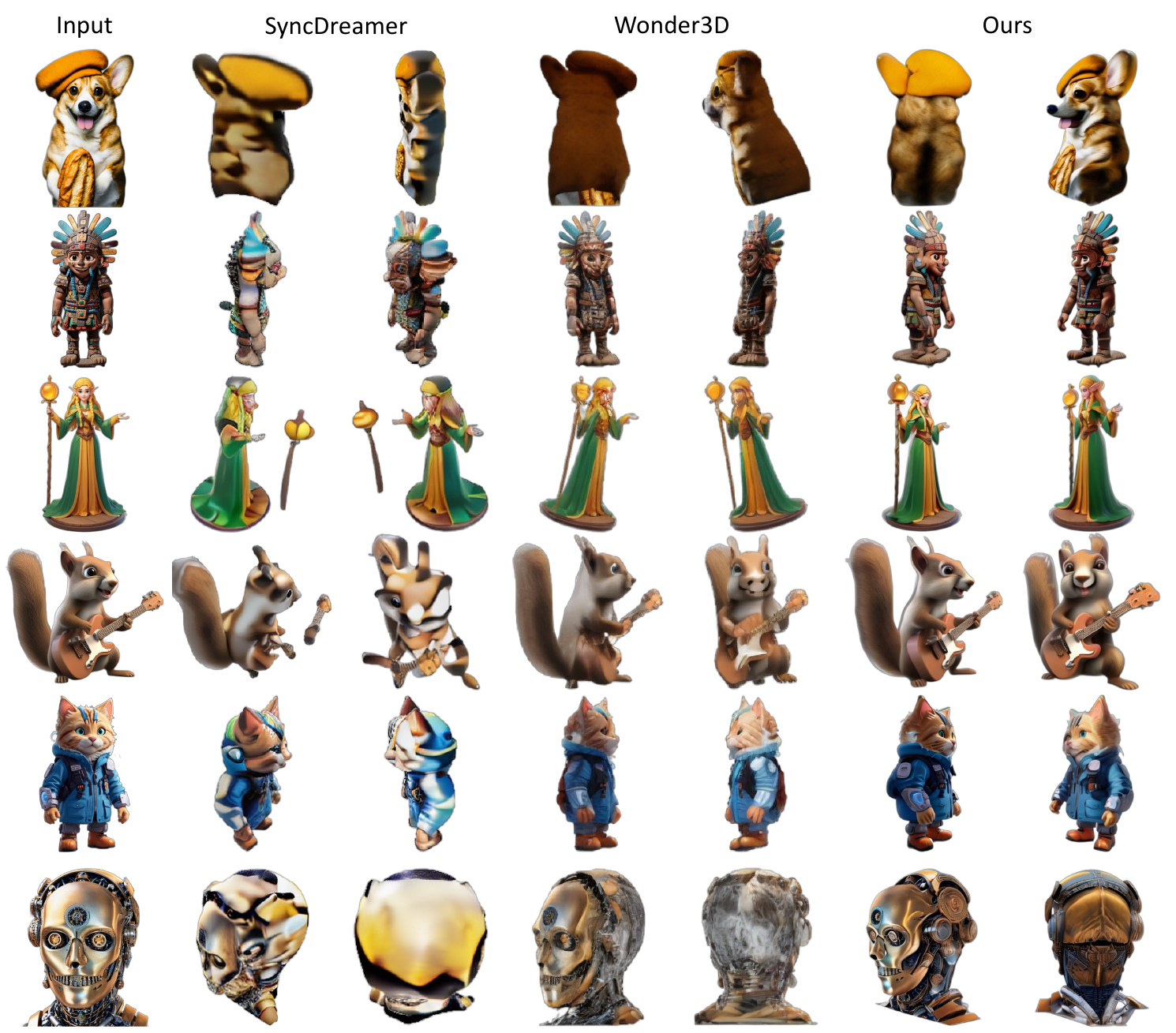}
    \caption{Qualitative comparison of generated views between the proposed \approach and state-of-the-art multi-view generation methods, including SyncDreamer~\cite{liu2023syncdreamer} and Wonder3D~\cite{long2023wonder3d}. The corresponding $360\degree$ videos are shown in the supplemental video.}
    \label{fig:mv_comparison}
    \vspace{-4mm}
\end{figure}

\subsection{Object-centric 3D Generation}
\subsubsection{Qualitative Comparisons.}
We evaluate the performance of the proposed \approach on image-to-3D and compare the results with state-of-the-art methods. The upper part of \cref{fig:gs_comparison} demonstrates qualitative comparisons between our approach with 3DGS-based methods TriplaneGaussian~\cite{zou2023triplane} and LGM~\cite{tang2024lgm}. Our approach demonstrates impressive quality improvement, while TriplaneGaussian and LGM tend to generate a much more blurry appearance due to the limited number of Gaussians they can produce. The lower part of \cref{fig:gs_comparison} presents a comparison between state-of-the-art SDS-based methods, including Magic123~\cite{Magic123} and ImageDream~\cite{wang2023imagedream}. \approach outperforms in both front-view alignment and fidelity. In contrast, objects generated by Magic123 exhibit inaccurate geometry and blurry backface, while ImageDream tends to generate assets with over-saturated textures. Importantly, our method achieves these results within 3 minutes, making significant acceleration compared to optimization-based methods which require more than 30 minutes.
\cref{fig:mv_comparison} illustrates comparisons on multi-view generation with SyncDreamer~\cite{liu2023syncdreamer} and Wonder3D~\cite{long2023wonder3d}, where our approach consistently achieves superior results with a higher resolution of 512 and much finer details. SyncDreamer tends to generate multi-view images that are simple in geometry and with over-saturated texture. We attribute this to the significant difference in structure between SyncDreamer and the original Stable Diffusion, thus failing to fully leverage the information from image pre-training. 
% As for Wonder3D, it produces images that are much less clear in texture, which we attribute to the powerful capabilities brought to our model by large-scale video pretraining.
As for Wonder3D, our model triples the number of generated multi-views and doubles the resolution, achieving significant improvements in both texture quality and multi-view consistency.

\subsubsection{Quantitative Comparisons.}
For qualitative comparisons, we conduct a user study to evaluate the generated 3D assets with human ratings. Specifically, we ask 58 volunteers to evaluate objects generated by \approach and other methods under 30 condition images by watching the rendered $360\degree$ spiral videos, and choose the best method in (a) Alignment: how well does the 3D asset align with the condition image; (b) Fidelity: how realistic is the generated object. \cref{tab:user_study} demonstrates the win rate of each method in the two criteria. \approach is regarded as the most convincing model of all, and it significantly outperforms other competitors in both image alignment and fidelity.

% \begin{table}[ht]
%   \caption{User study results.}
%   \label{tab:user_study}
%   \centering
  
% \end{table}

% \begin{table}[ht]
%   \caption{Quantitative comparison of dense multi-view synthesis on Google Scanned Objects dataset~\cite{GSO}.}
%   \label{tab:gso_nvs}
%   \centering
%   \begin{tabular}{@{}cccc@{}}
%     \toprule[2pt]
%     Approach & PSNR$\shortuparrow$ & SSIM$\shortuparrow$ & LPIPS$\shortdownarrow$\\
%     \midrule
%     RealFusion~\cite{melaskyriazi2023realfusion} & 15.26 & 0.722 & 0.283\\
%     Zero123~\cite{liu2023_zero1to3} & 18.93 & 0.779 & 0.166\\
%     SyncDreamer~\cite{liu2023syncdreamer}  & 20.05 & 0.798 & 0.146\\
%     MVDiffusion++~\cite{tang2024mvdiffusionpp} & 21.45 & 0.844 & 0.129\\
%     Ours & \textbf{22.38} & \textbf{0.892} & \textbf{0.102}\\
%   \bottomrule[2pt]
%   \end{tabular}
% \end{table}

\begin{table}[t]
\begin{minipage}[t]{0.48\textwidth}
% \begin{table}[t!]
    \centering
    \makeatletter\def\@captype{table}
    \caption{User study results. We collect 58 anonymous preference results on image alignment and fidelity.}
    \vspace{1mm}
    \resizebox{1\textwidth}{!}{
    \begin{tabular}{@{}ccc@{}}
    \toprule[2pt]
    Approach & Alignment & Fidelity\\
    \midrule
    SyncDreamer~\cite{liu2023syncdreamer}  & 10.34 & 5.17\\
    Wonder3D~\cite{long2023wonder3d} & 17.24 & 12.07\\
    TriplaneGaussian~\cite{zou2023triplane} & 6.90 & 3.45\\
    LGM~\cite{tang2024lgm} & 13.79 & 24.14\\
    Ours & \textbf{51.72} & \textbf{55.17}\\
  \bottomrule[2pt]
  \end{tabular}
    }
    \label{tab:user_study}
% \end{table}
\end{minipage}
\hfill
\begin{minipage}[t]{0.50\textwidth}
    \centering
    \makeatletter\def\@captype{table}
    \caption{Quantitative comparison of dense multi-view synthesis on Google Scanned Objects dataset~\cite{GSO}.}
    \vspace{1.0mm}
    \resizebox{1\textwidth}{!}{
    \begin{tabular}{@{}cccc@{}}
    \toprule[2pt]
    Approach & PSNR$\shortuparrow$ & SSIM$\shortuparrow$ & LPIPS$\shortdownarrow$\\
    \midrule
    RealFusion~\cite{melaskyriazi2023realfusion} & 15.26 & 0.722 & 0.283\\
    Zero-1-to-3~\cite{liu2023_zero1to3} & 18.93 & 0.779 & 0.166\\
    SyncDreamer~\cite{liu2023syncdreamer}  & 20.05 & 0.798 & 0.146\\
    MVDiffusion++~\cite{tang2024mvdiffusionpp} & 21.45 & 0.844 & 0.129\\
    Ours & \textbf{22.08} & \textbf{0.882} & \textbf{0.102}\\
  \bottomrule[2pt]
  \end{tabular}
    }
    \label{tab:gso_comparison}
\end{minipage}
\vspace{-4mm}
\end{table}

% maybe in the supplemental mat.
% \subsubsection{Diversity.} As a diffusion model, \approach is capable of generating multiple plausible results. As demonstrated in \cref{fig:diversity}, our approach generates a varity of reasonable novel views with different random seeds.

% \begin{figure}[ht]
%     \centering
%     \includegraphics[width=1.0\textwidth]{figs/comparison.pdf}
%     \caption{Our approach can generate diverse multi-view videos with an ambiguous single-view image as the condition.}
%     \label{fig:diversity}
% \end{figure}

\subsection{Novel View Synthesis}
For scene-level novel view synthesis, we test the performance of the proposed \approach on the 10-category subset of the CO3D dataset. To align with the settings of previous methods, we fine-tune \approach on videos in each category for only one epoch (denoted as ``fine-tuned''). The results are shown in \cref{tab:co3d-10}. Our approach achieves consistently better performance in terms of image metrics against previous state-of-the-art novel view synthesizers by a clear margin, demonstrating the effectiveness of utilizing pre-trained video diffusion models in scene-level novel view synthesis. Besides, \approach shows impressive generalization ability as the zero-shot version of \approach (only trained on MVImgNet) beats most of the competitors.

\subsubsection{Qualitative Comparisons.}
\begin{figure}[ht]
    \centering
    \includegraphics[width=1.0\textwidth]{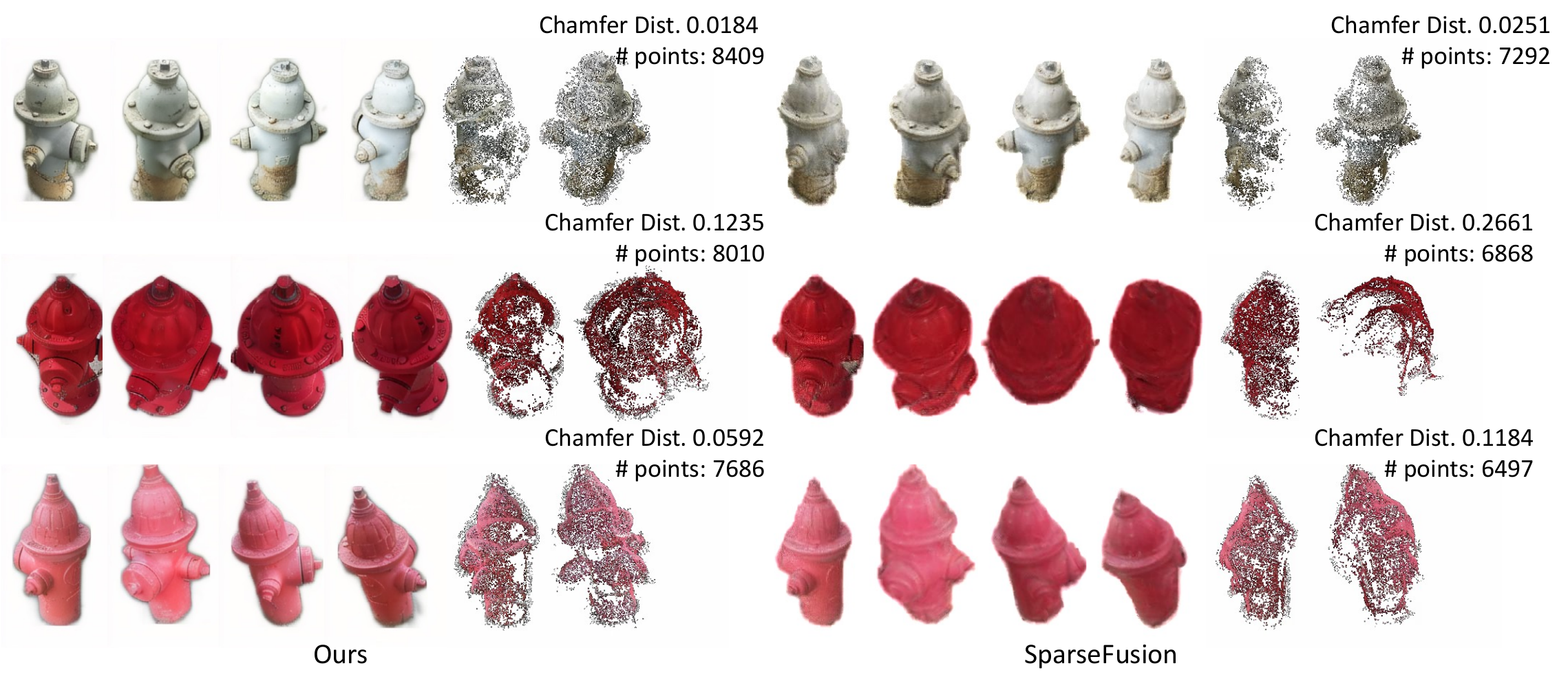}
    \vspace{-5mm}
    \caption{Qualitative comparison on scene-level novel view synthesis with SparseFusion on \textit{hydrant} subset of CO3D dataset. We reconstruct a sparse point cloud using COLMAP and report the number of points and Chamfer distance against the point cloud extracted with ground truth images.}
    \label{fig:scene}
    \vspace{-1.5mm}
\end{figure}

We show in \cref{fig:scene} qualitative comparisons on generated multi-views between SparseFusion~\cite{sparsefusion} on the \textit{hydrant} subset of the CO3D dataset~\cite{reizenstein21co3d}. For better comparison, we use COLMAP~\cite{schoenberger2016sfm, schoenberger2016mvs} to perform multi-view stereo reconstruction given camera poses and showcased, in \cref{fig:scene}, the number of points in the obtained point cloud and the Chamfer distance between the one reconstructed from ground truth images.
The point clouds reconstructed from \approach generated images contain more points and are closer to the point clouds reconstructed from real images. This indicates a significant advantage of our method in terms of both reconstruction quality and multi-view consistency.

%------------------------------------------------------------------------
\begin{table}[t!]
\footnotesize
\begin{center}

\caption{Quantitative comparison on 10-category subset of CO3D dataset~\cite{reizenstein21co3d}. We follow SparseFusion~\cite{sparsefusion}, benchmark view synthesis results with 2, 3, and 6 input views. The best result is highlighted in \textbf{bold}, while the second-best result is \underline{underscored}.}
\vspace{-1mm}
\label{tab:co3d-10}
\setlength{\tabcolsep}{2pt}
\resizebox{\linewidth}{!}{
\begin{tabular}{l ccc ccc ccc}
\toprule[2pt]
& \multicolumn{3}{c}{2 Views} & \multicolumn{3}{c}{3 Views} & \multicolumn{3}{c}{6 Views}\\
\cmidrule(r){2-4} \cmidrule(r){5-7} \cmidrule(r){8-10}
& PSNR$\shortuparrow$ & SSIM $\shortuparrow$ & LPIPS $\shortdownarrow$  & PSNR $\shortuparrow$ & SSIM $\shortuparrow$ & LPIPS $\shortdownarrow$ & PSNR $\shortuparrow$ & SSIM $\shortuparrow$ & LPIPS $\shortdownarrow$ \\
\midrule
PixelNeRF \cite{pixelnerf}       

& 19.52 & 0.667 & 0.327 & 20.67 & 0.712 & 0.293 & 22.47 & 0.776 & 0.241  \\

NerFormer \cite{reizenstein21co3d}       

& 17.88 & 0.598 & 0.382 & 18.54 & 0.618 & 0.367 & 19.99 & 0.661 &  0.332 \\

ViewFormer \cite{kulhanek2022viewformer}      

& 18.37 & 0.697 & 0.282 & 18.91 & 0.704 & 0.275 & 19.72 & 0.717 & 0.266  \\

EFT~\cite{He2020EpipolarT, sparsefusion}

&  20.85 & 0.680 & 0.289 &  \underline{22.71} & 0.747 & 0.262 & \underline{24.57} & 0.804 & 0.210  \\

VLDM~\cite{sparsefusion}               

& 19.55 & 0.711 &  0.247 & 20.85 & 0.737 &  0.225 & 22.35 & 0.768 &  0.201  \\

SparseFusion~\cite{sparsefusion}      

&  \underline{21.34} & \underline{0.752} & 0.225 &  {22.35} & 0.766 & 0.216 &  23.74 & 0.791 &  0.200  \\
\midrule

% \textbf{Ours} {\scriptsize(w. posed CLIP emb.)}

% & 19.67 & \underline{0.842} & \underline{0.113} & 22.04 & 0.855 &  0.095 &  \underline{23.74} & \underline{0.881} &  \textbf{0.074}  \\

% \textbf{Ours} {\scriptsize(w. Pl\"{u}cker emb.)}

% & 20.31 & \underline{0.842} & \underline{0.113} & 22.04 & 0.855 &  0.095 &  \underline{23.74} & \underline{0.881} &  \textbf{0.074}  \\

% \textbf{Ours} {\scriptsize(w/o. pre-training)}

% & 16.20 & 0.544 & 0.333 & 22.04 & 0.855 &  0.095 &  \underline{23.74} & \underline{0.881} &  \textbf{0.074}  \\

% \textbf{Ours} (zero-shot)            

% & 20.64 & \underline{0.734} & \underline{0.113} & 22.04 & 0.855 &  0.095 &  \underline{23.74} & \underline{0.881} &  \textbf{0.074}  \\
% \textbf{Ours} (fine-tuned)           

% &  \textbf{22.19} & \textbf{0.790} &  \textbf{0.093 } &  \textbf{23.31} & \textbf{0.860} &  \textbf{0.082} &  \textbf{24.70} & \textbf{0.890} &  \textbf{0.074}  \\
\textbf{Ours} {\scriptsize(w. camera emb.)} & 18.27 & 0.598 & 0.303 & 18.83 & 0.605 & 0.273 & - & - & - \\ 
\textbf{Ours} {\scriptsize(w. Pl\"{u}cker emb.)} & 20.11 & 0.714 & 0.277 & 20.39 & 0.727 & 0.223 & - & - & - \\ 
\textbf{Ours} {\scriptsize(w/o. pre-training)} & 16.20 & 0.544 & 0.333 & 16.76 & 0.576 & 0.303 & - & - & - \\ 
\textbf{Ours} (zero-shot) & 20.64 & 0.734 & \underline{0.213} & 22.04 & \underline{0.805} & \underline{0.141} & {23.94} & \underline{0.851} & \underline{0.126} \\ 
\textbf{Ours} (fine-tuned) & \textbf{22.19} & \textbf{0.790} & \textbf{0.093 } & \textbf{23.31} & \textbf{0.860} & \textbf{0.082} & \textbf{24.70} & \textbf{0.890} & \textbf{0.074} \\ 

\bottomrule[2pt]
\vspace{-3em}
\end{tabular}
}
\end{center}
\end{table}

\subsection{Ablations}
\begin{figure}[ht]
    \centering
    \includegraphics[width=1.0\textwidth]{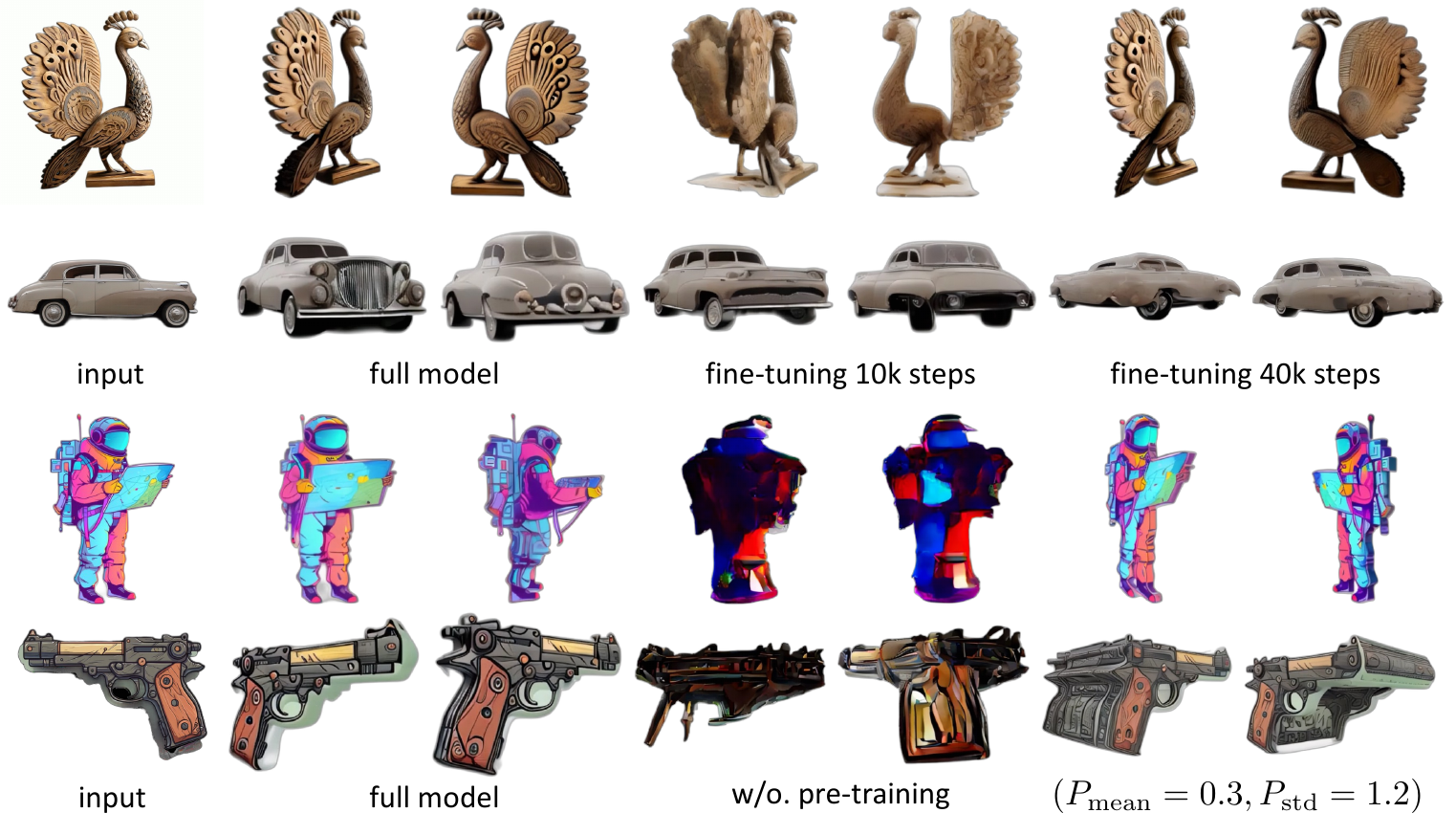}
    \vspace{-2mm}
    \caption{Ablations study. We show that the number of fine-tuning steps, large-scale pre-training, and stronger noise distribution are crucial for achieving promising results.}
    \label{fig:ablations}
    \vspace{-3mm}
\end{figure}

\label{sec:ablation}
\subsubsection{Pre-training.}
To demonstrate the impact of large-scale video pre-training, we trained a variant of the model from scratch on the constructed Objaverse $360\degree$ orbit video for 45k steps, which, apart from the initialized weights of the diffusion U-Net, is identical to the full model. As shown in \cref{fig:ablations}, this from-scratched variant completely failed to generate multi-view images even with doubled training steps compared to the full model, demonstrating the significant importance of adopting large-scale video pre-training.

\subsubsection{Number of Fine-tuning Steps.}
To illustrate the impact of the Objaverse dataset on the pre-trained model, we fine-tune the base model with different training steps, as shown in the top part of \cref{fig:ablations}. Due to the considerably lower complexity of renders from Objaverse objects, excessive fine-tuning steps can yield overly simplistic textures in the generated frames, while an inadequate number of fine-tuning steps may lead to strange and inaccurate geometries.

\subsubsection{Noise Distribution.}
To validate the necessity of adopting a large noise distribution during training, we fine-tune the base diffusion model with a smaller noise schedule ($P_{\text{mean}}=0.7$ and $P_{\text{std}}=1.6$), which is used in the image-to-video pre-training in Stable Video Diffusion. As demonstrated in the lower part of \cref{fig:ablations}, the model trained with a smaller noise distribution tends to generate geometric structures that are unreasonable (the pistol) or degenerated (the cartoon-style astronaut).
We attribute this phenomenon to the strong image signal we exploited and the relatively simple nature of the synthetic Objaverse dataset, indicating the need to shift the noise distribution towards stronger noise.

\subsubsection{Camera Conditioning.}
To assess the impact of the approach of camera conditioning, we introduce two variants that condition on camera poses by commonly used learnable camera embedding or the Pl\"{u}cker ray embedding. For a fair comparison, for both variants, we additionally concatenate the multi-view images used as conditions into the input of the diffusion U-Net and similarly set the weights of added channels to zero to minimize the gap. As shown in \cref{tab:co3d-10}, either camera embedding or Pl\"{u}cker embedding provides accurate 3D information and results in a significant degeneration in model performance.

\section{Limitations and Conclusion}
\subsubsection{Limitations.}
Although achieves state-of-the-art performance in 3D generation, \approach would produce unsatisfying results for some complex objects or scenes, such as inconsistency among multiple views or unreasonable geometries. Concrete failure cases and analyses are discussed in the supplemental material.
\subsubsection{Conclusion.}
% In this paper, we propose \approach, a novel method for generating consistent multi-view images and novel views with image-to-video diffusion models for object-centric and scene-level scenarios. By fine-tuning the base video diffusion model on 3D datasets, we extend video diffusion models to multi-view consistent multi-view 3D generators. Specifically, for object-centric image-to-3D generation, we fine-tune the video diffusion for synthesizing 360-degree videos of synthetic objects and restore the underlying 3D asset using Gaussian splatting with tailored initialization and perceptual loss. For scene-level novel view synthesis, we extend the video diffusion model to condition on a user-specified camera path with a PixelNeRF encoder for generating novel views in real-world video clips. We conduct extensive experiments to validate the effectiveness of the proposed approach, illustrating its remarkable performance in generating multi-view consistent novel views and generalization ability. We hope our method can serve as an efficient and powerful approach for high-quality image-to-3D and novel view synthesis and could pave the way for more extensive applications of video diffusion models in 3D generation.
In this paper, we propose \approach, a novel method for generating consistent multi-view images with image-to-video diffusion models. By fine-tuning the base video diffusion model on 3D datasets, we extend video diffusion models to effective 3D generators. Specifically, for object-centric 3D generation, we fine-tune the video diffusion on synthesizing $360\degree$ videos of 3D objects to predict dense views given a single image. To obtain the underlying 3D asset from generated views, we propose a tailored reconstruction pipeline with designed initialization and texture refinement, enabling the reconstruction of high-quality 3D Gaussians or delicate textured meshes within 3 minutes. 
We further extend our framework to scene-level novel view synthesis, achieving precise control over the camera path with great multi-view consistency.
We conduct extensive experiments to validate the effectiveness of the proposed approach, illustrating its remarkable performance in generating consistent multi-views and generalization ability. We hope our method can serve as an efficient and powerful approach for high-quality 3D generation and could pave the way for more extensive applications of video diffusion models in 3D tasks.

\clearpage  % TODO REVIEW/FINAL: This \clearpage needs to be removed from both review and camera-ready versions.

\bibliographystyle{splncs04}
\bibliography{main}
\end{document}